\documentclass[runningheads]{llncs}

\usepackage[T1]{fontenc}
\usepackage{graphicx}
\usepackage{booktabs}
\usepackage{amsmath}
\usepackage{amssymb}
\usepackage{amsfonts}
\usepackage{hyperref}

\newcommand{\runinhead}[1]{\par\medskip\noindent\textbf{#1}\hspace{0.5em}}

\usepackage{xcolor}
\usepackage[table]{xcolor}
\usepackage{multirow}
\usepackage{multicol}

\usepackage{xspace}
\makeatletter
\DeclareRobustCommand\onedot{\futurelet\@let@token\@onedot}
\def\@onedot{\ifx\@let@token.\else.\null\fi\xspace}

\def\etal{\emph{et al}\onedot}
\makeatother

\makeatletter
\newcommand{\thickhline}{%
\noalign {\ifnum 0=`}\fi \hrule height 1pt
\futurelet \reserved@a \@xhline
}
\makeatother

\usepackage{amsmath,amsfonts,bm}

\def\eqref#1{equation~\ref{#1}}
\def\1{\bm{1}}

\def\vp{{\bm{p}}}

\def\mH{{\bm{H}}}

\def\mR{{\bm{R}}}

\def\mV{{\bm{V}}}

\DeclareMathAlphabet{\mathsfit}{\encodingdefault}{\sfdefault}{m}{sl}
\SetMathAlphabet{\mathsfit}{bold}{\encodingdefault}{\sfdefault}{bx}{n}

\newcommand{\R}{\mathbb{R}}

\usepackage{pifont}
\definecolor{softgreen}{RGB}{46,125,50}
\definecolor{softred}{RGB}{198,40,40}
\newcommand{\cmark}{\textcolor{softgreen}{\ding{51}}}

\usepackage{orcidlink}
\usepackage{marvosym}

\begin{document}

\title{Self-supervised Learning Matters: A Simple Ensemble Solution for Micro-Gesture Recognition}
\titlerunning{Self-supervised Learning Matters}

% \author{Anonymous}
% \institute{Anonymous}

\author{Tingyi Liu\inst{1}\orcidlink{0009-0006-2486-7516} \and 
Kun Li\inst{2}\textsuperscript{(\Letter)}\orcidlink{0000-0001-5083-2145}  \and 
Fei Wang\inst{1,3,4}\orcidlink{0009-0004-1142-6434}
\and
Junjie Chen\inst{1,3}\orcidlink{0009-0001-5288-048X}
\and 
Zhiliang Wu \inst{5}\orcidlink{0000-0002-6597-8048}
\and
Jihao Gu\inst{6}\orcidlink{0009-0009-0141-4807} \and 
Haixu Liu\inst{7,8}\orcidlink{0009-0007-8115-0826}
\and 
Dan Guo\inst{1,3}\orcidlink{0000-0003-2594-254X}
}

\authorrunning{Tingyi Liu et al.}

\institute{
Hefei University of Technology, Hefei, China\and
United Arab Emirates University, Al Ain, United Arab Emirates 
\email{kunli.hfut@gmail.com}
\and
Institute of Artificial Intelligence, Hefei Comprehensive National Science Center, Hefei, China\and
Anhui Evolution Technology Co., Ltd., Hefei, China \and
Nanyang Technological University, Singapore \and 
University College London, United Kingdom \and
The University of Sydney, Sydney, New South Wales, Australia \and
Beijing QBoson Quantum Technology Co., Ltd., China
}

\maketitle

\begin{abstract}
In this paper, we present XInsight Lab's solution to the micro-gesture classification track of the 4th MiGA Challenge at IJCAI 2026, in which our solution ranked first and achieved a new state-of-the-art result. 
We propose a multimodal ensemble framework that integrates a self-supervised RGB-based model with supervised multi-stream models from previous solutions. The self-supervised RGB model is pretrained on 120K unlabeled clips via masked video modeling and then fine-tuned on iMiGUE. This simple yet effective RGB baseline achieves 69.224\% top-1 accuracy on the iMiGUE test set, demonstrating the benefit of learning transferable representations from unlabeled in-domain videos. By incorporating this model as a complementary branch, the final ensemble reaches 74.419\% top-1 accuracy, surpassing the previous state of the art by 1.206 percentage points. Experimental results on iMiGUE, including ablation studies on the ensemble strategy, validate the effectiveness of self-supervised RGB representation learning for micro-gesture recognition.
\end{abstract}

\section{Introduction}
\label{sec:intro}

Micro-gestures (MGs)~\cite{liu2021imigue,chen2023smg,wang2026imigue} are subtle, spontaneous body movements that may reflect hidden cognitive or affective states~\cite{wang2026gait,wang2026xinsight}, such as stress, hesitation, or emotional suppression. Typical examples include brief forehead touches, lip covering, or repetitive finger tapping in high-pressure interactions. 
Micro-gesture recognition~\cite{liu2021imigue,chen2023smg,wang2026imigue,wang2026micro} is particularly challenging compared with general action recognition~\cite{li2023data,li2023datae,hao2022group,hao2022attention,wang2025exploiting}, as MG instances are brief, spatially subtle, and often obscured by background motion, pose-estimation noise, and large intra-class variation. 

In micro-gesture recognition, previous work can be roughly grouped into skeleton-based methods~\cite{li2023joint,huang2023micro} and multi-stream-based methods~\cite{chen2024prototype,gu2025mm}. Skeleton-based methods adopt GCN, 3D-CNN~\cite{li2023joint}, and transformer backbones~\cite{huang2023micro} to model the subtle limb dynamics of MGs. 
Recent methods have further advanced micro-gesture recognition through multi-stream fusion. For instance, Chen~\etal~\cite{chen2024prototype} proposed a prototype-calibrated framework that jointly models prototypes from skeleton and RGB modalities. Gu~\etal~\cite{gu2025mm} further explored large-scale multimodal fusion by integrating skeleton, RGB, depth, optical flow, and Taylor-series representations.
Despite their effectiveness, these methods rely heavily on supervised annotations, while the potential of self-supervised pretraining for micro-gesture recognition remains largely unexplored.

Self-supervised video pretraining can learn transferable representations from unlabeled data, making it well-suited to scarce and imbalanced micro-gesture annotations. 
Recently, Wang~\etal~\cite{wang2026imigue} introduced iMiGUE-3K, a large-scale micro-gesture dataset, together with MG-FM-RGB, an RGB-based self-supervised framework~\cite{thoker2025smile} that demonstrates the potential of self-supervised learning for micro-gesture recognition. Inspired by this work, we adopt MG-FM-RGB as a self-supervised video model for micro-gesture recognition. Specifically, the encoder is first pretrained on 120K unlabeled samples via masked video modeling and then transferred to iMiGUE for fine-tuning. Finally, we integrate this model into the previous MM-Gesture framework~\cite{gu2025mm} as a complementary RGB stream for multimodal ensemble learning.

The contributions of this paper are as follows:
\begin{itemize}
\item We adapt a simple yet effective self-supervised RGB video framework based on SMILE masked video modeling for micro-gesture recognition. The framework leverages in-domain unlabeled videos to learn transferable RGB representations without action-level annotations.

\item We integrate the self-supervised RGB stream with the existing MM-Gesture multimodal framework, enabling complementary fusion between learned RGB representations and other modality-specific cues. This combined solution achieves 74.419\% top-1 accuracy on the iMiGUE test set and ranks first in Track~1 of the 4th MiGA Challenge.
\end{itemize}

%%%%%%%%%%%%%%%%%%%%%%%%%%%%%%%%%%%%%%%
\section{Related Work}
\label{sec:related}

\subsection{Micro-Gesture Analysis}
Micro-gesture analysis~\cite{chen2023smg,liu2021imigue} has recently attracted growing research interest, driven by the emergence of dedicated benchmarks~\cite{guo2024benchmarking,li2026bench,li2025mmad,wang2026imigue}, advanced modeling approaches~\cite{gu2025motion,li2025prototypical}, and community competitions~\cite{chen20253rd,guo2024mac,li2025mac}. Micro-gesture analysis covers a range of tasks, including micro-gesture recognition~\cite{chen2024prototype,gu2025mm,li2023joint,shang2025cross}, online micro-gesture detection~\cite{liu2025online,liu2024micro}, and behavior-based emotion understanding~\cite{wang2026weak,xia2025hybrid}.

Micro-gesture recognition in the MiGA series competitions has attracted a wide range of approaches from participating teams. 
Li~\etal~\cite{li2023joint} combine PoseConv3D~\cite{duan2022revisiting} with a joint skeletal–semantic embedding loss. Chen~\etal~\cite{chen2024prototype} augment PoseConv3D with a prototype-refinement module that calibrates ambiguous samples on cross-modal fused features. Huang~\etal~\cite{huang2024multi} couple Res2Net3D with an EHCT branch into a heterogeneous skeleton+RGB ensemble. Wang~\etal~\cite{wang2024multimodal} build a CLIP-based RGB stream with video-level distillation and inject CLIP text embeddings into the PoseConv3D skeleton stream. MM-Gesture~\cite{gu2025mm} fuses six streams, including joint, limb, RGB, Taylor video, optical flow, and depth, into the previous best-performing method on iMiGUE. 
In contrast, Hu~\etal~\cite{hu2025enhancing} augment a ViT-Base RGB backbone with a global-aware importance module that focuses on fine-grained foreground regions. 
However, all of these methods are trained purely with supervised labels on iMiGUE, SMG, or MA-52, while the potential of self-supervised pretraining on unlabeled data remains largely unexplored.

\subsection{Self-supervised Learning}
Self-supervised learning~\cite{lin2020ms2l,li20213d,guo2022contrastive,he2022masked,mao2023masked} learns representations from unlabeled data through pretext tasks, without manual annotations. 
In video understanding, MAE~\cite{he2022masked} reconstructs randomly masked image patches and achieves competitive transfer to downstream tasks. VideoMAE~\cite{tong2022videomae} extends masked modeling to video using tube masking at high ratios. 
VideoMAEv2~\cite{wang2023videomae} scales masked video autoencoders with dual masking, lowering pretraining cost while preserving reconstruction quality and improving transfer to downstream video understanding tasks. 
More recent methods reconstruct features from a frozen pretrained encoder instead of raw pixels, yielding semantically richer targets.  
SMILE~\cite{thoker2025smile} combines feature-target reconstruction with synthetic motion injection for video pretraining. 
Although it has demonstrated effectiveness on general action benchmarks, masked video modeling remains largely unexplored for micro-gesture recognition, where in-domain unlabeled videos are available but labeled annotations are limited. 
To address this gap, Wang~\etal~\cite{wang2026imigue} introduced MG-FM-RGB, a SMILE-based framework pretrained on iMiGUE-3K-u and fine-tuned on iMiGUE. We therefore adopt MG-FM-RGB as the baseline model in this work.

%%%%%%%%%%%%%%%%%%%%%%%%%%%%%%%%%%%%
\section{Methodology}
\label{sec:method}

\subsection{Network Architecture}
\label{sec:method:overview}

\begin{figure}[t]
\centering
\includegraphics[width=\linewidth]{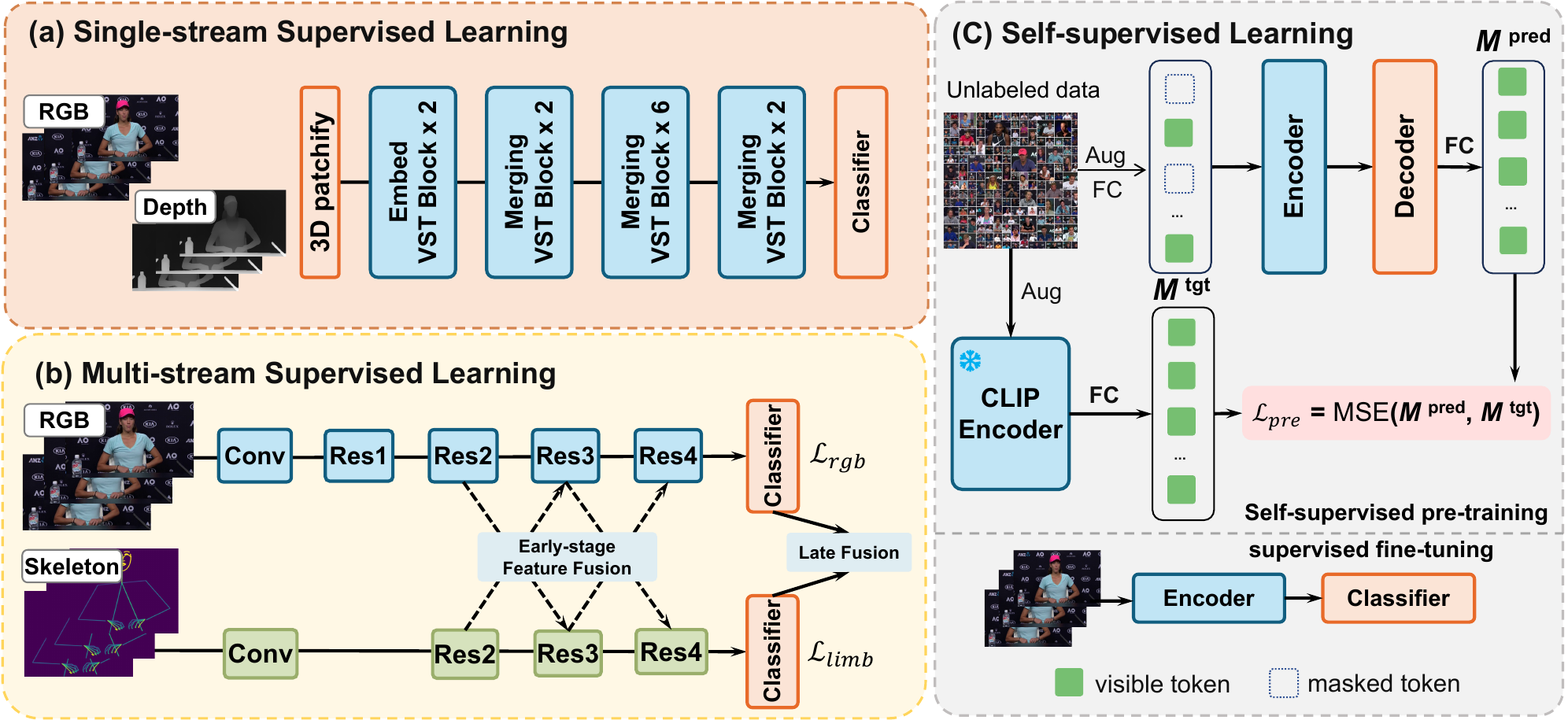}
\caption{Overview of the proposed multi-stream ensemble solution for micro-gesture recognition.
(a) Single-stream supervised learning with Video Swin Transformer~\cite{liu2022video} using RGB and depth modalities as inputs.
(b) Multi-stream supervised learning with PoseConv3D~\cite{duan2022revisiting} to learn complementary representations from RGB and skeleton data.
(c) Self-supervised learning with MG-FM-RGB~\cite{wang2026imigue}, which is built upon SMILE~\cite{thoker2025smile}.
}
\label{fig:pipeline}
\end{figure}

As shown in Figure~\ref{fig:pipeline}, our solution integrates three types of models with complementary input modalities. Given an input video sample, the goal is to predict its category label $y \in \{1, \dots, C\}$ among the $C{=}32$ iMiGUE classes. The sample is processed by $K{=}5$ streams built upon three representative backbones. For the single-stream supervised learning paradigm, we feed RGB and depth inputs into Video Swin Transformer~\cite{liu2022video}. 
For the multi-stream supervised learning paradigm, we employ PoseConv3D~\cite{duan2022revisiting} with skeleton and RGB inputs. 
For the self-supervised paradigm, we use MG-FM-RGB~\cite{wang2026imigue}, which is pretrained on unlabeled in-domain videos. Each stream $f_k$ outputs a softmax distribution $\vp_k=(p_{k,1},\dots,p_{k,C}) \in \R^{C}$, and the final prediction is obtained by weighted average fusion:
\begin{equation}
\vp = \sum_{k=1}^{K} w_{k}\vp_{k}, \qquad \hat{y} = \arg\max_{c\in\{1,\dots,C\}} p_c,
\end{equation}
where $\vp=(p_1,\dots,p_C)$, and $\{w_k\}_{k=1}^{K}$ denote the ensemble weights, with $w_k\ge0$ and $\sum_{k=1}^{K}w_k=1$.

\subsection{Single-stream Supervised Learning}
\label{sec:method:swin}
Following MM-Gesture~\cite{gu2025mm}, we employ Video Swin Transformer~\cite{liu2022video} as the baseline model and train it with RGB and depth modalities. Video Swin Transformer extends the shifted-window self-attention mechanism of Swin Transformer~\cite{liu2021swin} to 3D spatiotemporal windows, enabling the model to capture local temporal dynamics together with spatial patterns. Given an input clip, the model first splits it into 3D patch tokens. It then processes the tokens through four hierarchical stages, where regular and shifted 3D window attention are alternately applied, and patch merging layers reduce the spatial resolution between stages. We use the small variant, Swin-S, followed by global pooling and a linear classification head to predict the 32 micro-gesture categories.

The RGB stream takes the raw clip $\mV_{rgb} \in \R^{T \times H \times W \times 3}$ as input. It is pretrained on the Micro-Action-52 (MA-52) dataset~\cite{guo2024benchmarking} and fine-tuned on iMiGUE. 
The depth stream takes a depth clip $\mV_{depth} \in \R^{T \times H \times W \times 1}$ as input. Each depth frame is generated from the corresponding RGB frame by a monocular depth estimator. Unlike the RGB stream, the depth stream is trained from scratch on iMiGUE. Let $\vp_m=(p_{m,1},\dots,p_{m,C})$ denote the softmax output of modality $m$. Both streams are optimized using the cross-entropy loss:
\begin{equation}
\mathcal{L}_{m} = \mathrm{CE}(\vp_m, y) = -\log p_{m,y}, \quad m \in \{\mathrm{rgb}, \mathrm{depth}\},
\end{equation} 
where $p_{m,y}$ is the probability assigned to the ground-truth class by modality $m$. Finally, the softmax outputs of these models are used as separate streams in the final ensemble.

\subsection{Multi-stream Supervised Learning}
\label{sec:method:posec3d}
Here, we adopt PoseConv3D~\cite{duan2022revisiting} to model RGB and skeleton data, which has shown strong performance in previous micro-gesture recognition solutions~\cite{chen2024prototype,gu2025mm,xu2025towards}. 
PoseConv3D encodes pose as heatmap volumes and fuses them with RGB at the decision level. 
Specifically, we first convert the 2D skeletal keypoints into heatmap representations: for each frame, every limb is drawn as a Gaussian map along the segment between its two endpoint joints, and the maps are stacked over time into a 3D limb-heatmap volume
\begin{equation}
\mH_{L} \in \R^{T \times H \times W \times E},
\end{equation}
where $T$ is the number of frames, $H \times W$ the spatial resolution, and $E$ the number of skeletal limbs. The RGB clip $\mR \in \R^{T \times H \times W \times 3}$ and the limb volume $\mH_{L}$ are each encoded by a PoseConv3D backbone into a modality-specific softmax distribution $\vp_m=(p_{m,1},\dots,p_{m,C})$, $m \in \{\mathrm{R}, \mathrm{L}\}$, trained with a cross-entropy loss $\mathcal{L}_{m} = \mathrm{CE}(\vp_m, y)=-\log p_{m,y}$.

We build two streams from these backbones. The limb stream uses the model trained on $\mH_{L}$ alone. The skeleton+RGB stream couples the limb and RGB backbones: we fine-tune them jointly under the summed loss
\begin{equation}
\mathcal{L}_{R+L} = \mathcal{L}_{R} + \mathcal{L}_{L},
\end{equation}
and average their softmax probabilities $\vp_{R}$ and $\vp_{L}$ at inference,
\begin{equation}
\vp_{R+L} = \tfrac{1}{2}\left(\vp_{R} + \vp_{L}\right).
\end{equation}
Both streams are trained on iMiGUE for 32-class micro-gesture classification.

\subsection{Self-supervised Learning}
\label{sec:method:smile}
To learn transferable representations from large-scale unlabeled videos without manual annotations, we adopt SMILE-based MG-FM-RGB~\cite{wang2026imigue,thoker2025smile} as the self-supervised RGB stream. MG-FM-RGB is pretrained on unlabeled iMiGUE-3K~\cite{wang2026imigue} clips via masked video modeling and then fine-tuned on iMiGUE. Specifically, each clip is divided into $N$ spatiotemporal tubelets, and a subset $\mathcal{M}$ is randomly masked. The visible tubelets are encoded by a transformer encoder $\mathrm{Enc}_{\theta}$, while a lightweight decoder $\mathrm{Dec}_{\phi}$ reconstructs the masked tubelets in the feature space of a frozen CLIP encoder $g_{\psi}$, rather than in pixel space:
\begin{equation}
\mathcal{L}_{pre} =
\frac{1}{|\mathcal{M}|}
\sum_{i \in \mathcal{M}}
\left\|
\left[
\mathrm{Dec}_{\phi} \left(\mathrm{Enc}_{\theta}(\mV_{vis})\right)
\right]_i
-
\left[
g_{\psi}(\mV)
\right]_i
\right\|_{2}^{2}.
\end{equation}
Here, $[\cdot]_i$ denotes the feature aligned with the $i$-th masked tubelet. To enhance the sparse temporal cues in micro-gesture videos, SMILE further injects synthetic motion into the clip and masks tubelets along the inserted motion trajectory. After pretraining, we fine-tune the encoder for micro-gesture classification. Specifically, we retain $\mathrm{Enc}_{\theta}$, discard the decoder, and attach a linear classification head to the mean-pooled token features. The encoder and classification head are optimized on iMiGUE using the cross-entropy loss, and the resulting softmax output is used as one stream in the final ensemble. 

\section{Experiments}
\label{sec:experiments}

\subsection{Experimental Setup}
\label{sec:exp:setup}

\runinhead{Datasets.}
iMiGUE~\cite{liu2021imigue} is a benchmark for spontaneous micro-gesture recognition, containing 32 categories: 31 micro-gesture classes and one non-MG/background class. It provides training, validation, and test splits with 12,893, 777, and 4,562 clips and exhibits a highly long-tailed distribution, where the number of training samples per class ranges from 1 to 3,765. iMiGUE-3K~\cite{wang2026imigue} is a large-scale in-the-wild micro-gesture dataset collected from press-interview videos. MG-FM-RGB was pretrained on iMiGUE-3K-u, a subset of iMiGUE-3K containing 120K unlabeled clips for self-supervised pretraining. 
MA-52~\cite{guo2024benchmarking} is a large-scale whole-body micro-action recognition benchmark that contains 22,422 video instances across 52 action-level categories.

\runinhead{Evaluation metrics.}
We report top-1 classification accuracy on the iMiGUE test split, the official evaluation metric of the MiGA challenge series.

\runinhead{Implementation details.}
Pose keypoints are extracted with OpenPose and reduced to 36 upper-body, hand, and face keypoints, from which the limb representation is formed. A per-frame depth map is estimated from the RGB video. The Video Swin RGB backbone is pretrained on MA-52~\cite{guo2024benchmarking} and fine-tuned on iMiGUE, while the depth backbone is trained from scratch on iMiGUE. For the SMILE stream, we fine-tune the pretrained ViT-Base encoder on iMiGUE for 50 epochs with a batch size of 16, using AdamW with a peak learning rate of $5\times10^{-5}$, weight decay 0.05, layer-wise decay 0.75, a drop-path rate of 0.1, and a 5-epoch warmup. Each clip is sampled to 16 frames at a temporal stride of 2 and resized to $224\times224$, with mixup, cutmix, and label smoothing disabled. At inference time, we average predictions over 21 augmented views. The five streams are fused by a weighted average of their softmax probabilities, with weights tuned on the validation set.

\subsection{Experimental Results}
\label{sec:results}

We compare our method with representative approaches from prior MiGA Track~1 submissions. Table~\ref{tab:single-branch} summarizes the results of HFUT-VUT~\cite{li2023joint,chen2024prototype}, EHCT~\cite{huang2023micro}, MM-Gesture~\cite{gu2025mm}, and our solution. Our final ensemble combines PoseConv3D, Video Swin Transformer, and SMILE-based MG-FM-RGB streams, achieving 74.419\% top-1 accuracy on the iMiGUE test split and ranking first in the 2026 edition. Compared with the previous best result of 73.213\% from the MiGA'25 winner, our method brings a 1.206 percentage-point improvement.

As shown in Table~\ref{tab:ensemble}, the self-supervised RGB stream MG-FM-RGB alone achieves 69.224\% top-1 accuracy. The four-stream ensemble without the MG-FM-RGB obtains 72.591\%, and incorporating the self-supervised stream MG-FM-RGB raises the accuracy to 74.419\%. These results demonstrate that the self-supervised RGB representation provides complementary information to the supervised multimodal ensemble.

\begin{table}[t]
\centering
\caption{Top-3 submissions of the past four MiGA Track-1 editions (2023--2026) on the iMiGUE test split, ranked by top-1 accuracy. Modalities: $\mathbf{J}$=joint, $\mathbf{L}$=limb, $\mathbf{R}$=RGB, $\mathbf{T}$=Taylor video, $\mathbf{F}$=optical flow, $\mathbf{D}$=depth video, $\mathbf{H}$=heatmap.}
\label{tab:single-branch}
\resizebox{1.0\linewidth}{!}{
\begin{tabular}{|c|c|c|c|c|}
\hline\thickhline 
\rowcolor{gray!25} {Rank} & {Team} & {Core Methodology} & {Modality} & {Acc. (\%)} \\ 
\hline
MiGA'23 1st & HFUT-VUT~\cite{li2023joint} & PoseConv3D & $\mathbf{J+L}$ & 64.12 \\
MiGA'23 2nd & NPU-Stanford~\cite{huang2023micro} & Hyperformer & $\mathbf{J}$ & 63.02 \\ 
\hline
MiGA'24 1st & HFUT-VUT~\cite{chen2024prototype} & PoseConv3D & $\mathbf{J+L+R}$ & 70.254 \\
MiGA'24 2nd & NPU-MUCIS~\cite{huang2024multi} & Res2Net3D+GCN & $\mathbf{J+R}$ & 70.188 \\
MiGA'24 3rd & ywww11~\cite{wang2024multimodal} & PoseConv3D+CLIP & $\mathbf{J+R}$ & 68.917 \\ 
\hline
MiGA'25 1st & HFUT-VUT~\cite{gu2025mm} & PoseConv3D+VideoSwinT & $\mathbf{J+L+R+T+F+D}$ & 73.213  \\
MiGA'25 2nd & awuniverse~\cite{hu2025enhancing} & ViT-Base+GAIE & $\mathbf{R}$ & 68.697 \\
MiGA'25 3rd & Lonelysheep~\cite{xu2025towards} & PoseConv3D & $\mathbf{J+L}$  & 67.010 \\
\hline
\rowcolor{gray!10}  MiGA'26 1st & \textbf{XInsight Lab (Ours)} & PoseConv3D+VideoSwinT+MG-FM-RGB & $\mathbf{J+L+R+D}$ & \textbf{74.419}  \\
MiGA'26 2nd & YUV & - & - & 74.243\\
MiGA'26 3rd & xd & - & -  & 71.372 \\  
MiGA'26 4th & AIM~\cite{zhang2026amulti} & Decoupled ST-CNN + PoseC3D + Swin3D + R(2+1)D & $\mathbf{J+H+R}$  & 68.128 \\ \hline
\end{tabular}}
\end{table}

\begin{table}[t]
\centering
\caption{Ensemble accuracy on iMiGUE test set under different models: PoseC3D~\cite{duan2022revisiting}, Video Swin Transformer~\cite{liu2022video}, MG-FM-RGB$^\ast$ denotes our implementation results using official code.}
\tabcolsep 4pt
\resizebox{1.0\linewidth}{!}{
\begin{tabular}{|c|cccc|c|c|}
\hline\thickhline
\rowcolor{gray!25} Model & PoseC3D-L & PoseC3D-LR & VideoSwinT & VideoSwinT & MG-FM-RGB$^\ast$  & Acc. (\%) \\
\hline 
Modality & Limb & Limb + RGB & RGB & Depth & RGB  & \\
Pre-training & iMiGUE & iMiGUE & MA-52 & N/A & iMiGUE-3K-u  & \\ 
Fine-tuning & iMiGUE & iMiGUE & iMiGUE & iMiGUE & iMiGUE &   \\ \hline
\multirow{6}{*}{Single} 
& \cmark  & & & & & 64.686 \\
& & \cmark & & & &  68.917 \\
& & & \cmark & & &  66.615 \\
& & & & \cmark & &  65.212 \\
& & & &  & \cmark   & 69.224 \\
\hline
& \cmark & \cmark & \cmark  & \cmark  & & 72.591 \\
\multirow{-2}{*}{Ensemble}
& \cmark & \cmark & \cmark &\cmark & \cmark & \textbf{74.419} \\
\hline
\end{tabular}
}
\label{tab:ensemble}
\end{table}

Finally, we analyze the multimodal ensemble strategy. As shown in Table~\ref{tab:ensemble}, we construct an ensemble using multiple models, including PoseConv3D~\cite{duan2022revisiting} and Video Swin Transformer~\cite{liu2022video} with the modalities adopted in MM-Gesture~\cite{gu2025mm}, and MG-FM-RGB~\cite{wang2026imigue}, a SMILE-based self-supervised RGB model that we fine-tune on iMiGUE. By incorporating this self-supervised stream into the existing ensemble, the top-1 accuracy improves from 72.591\% to 74.419\%. These results indicate that the proposed RGB-based self-supervised baseline provides complementary representations to the existing RGB, depth, and limb-based streams, thereby further enhancing the overall ensemble performance.

\section{Conclusion}
\label{sec:conclusion}

In this paper, we presented a simple multi-stream ensemble for micro-gesture recognition. The streams are independently trained or fine-tuned under distinct training regimes, ranging from supervised pretraining for skeleton and video streams to self-supervised masked video modeling on iMiGUE-3K-u; their predictions are fused by a weighted softmax average. On the iMiGUE test split, the strongest single stream reaches a top-1 accuracy of 69.224\%, while the full ensemble reaches 74.419\%, ranking first in Track~1 of the 4th MiGA Challenge at IJCAI~2026. 

In future work, we plan to design micro-gesture-specific self-supervised pretraining with motion magnification~\cite{wang2024eulermormer,wang2024frequency} to better capture the short-duration and low-amplitude motion of micro-gestures, and to explore semantic priors from multimodal large language models for ambiguous and long-tailed categories.

\section*{Acknowledgments}
This work was supported by Anhui Provincial Natural Science Foundation (2408085J040), National Key R\&D Program of China (2024YFB3311600), National Natural Science Foundation of China (62272144, 72188101), the Major Project of Anhui Provincial Science and Technology Breakthrough Program (202423k09020001), and the New Cornerstone Science Foundation through the XPLORER PRIZE.

\bibliographystyle{splncs04}
\bibliography{ref}

\end{document}